\documentclass[letterpaper, 10 pt, journal, twoside]{ieeetran} 
\usepackage{newfloat}
\usepackage{caption}
\usepackage{hyperref,breakurl}
\DeclareFloatingEnvironment[fileext=cmh,placement={!ht},name=List]{myfloat}
\usepackage{ifpdf}
\usepackage{amsfonts}
\usepackage{amssymb}
\usepackage[cmex10]{amsmath}
\usepackage{multirow}
\usepackage{tikz}

\usepackage{hyperref,breakurl}

\usepackage{url}
\usetikzlibrary{decorations.markings,arrows}
\usetikzlibrary{decorations.shapes}
\tikzset{decorate sep/.style 2 args=
 {decorate,decoration={shape backgrounds,shape=circle,shape size=#1,shape sep=#2}}}
\usetikzlibrary{plotmarks}
\usetikzlibrary{shadings}
\usepackage{graphicx}
\usepackage{float}
\usepackage{amscd,amsgen,amsfonts,amsbsy}
\usepackage{mathrsfs}
\usepackage{color}
\usepackage{textcomp}
\usepackage{latexsym,graphicx}
\usepackage[british]{babel}
\usepackage[font=small,labelfont=bf]{caption}
\usepackage{pgf}
\usepackage{multirow}
\usepackage{nicefrac}
\usepackage{algorithm}
\usepackage{algorithmicx}

\usepackage[noend]{algpseudocode}
\algnewcommand{\algorithmicgoto}{\textbf{go to}}%
\algnewcommand{\Goto}[1]{\algorithmicgoto~\ref{#1}}%
\algnewcommand{\algoand}{\textbf{and }}
\algnewcommand{\algoor}{\textbf{or }}
\algnewcommand{\algotrue}{\textbf{True }}
\algnewcommand{\algofalse}{\textbf{False }}

\IEEEoverridecommandlockouts
\usepackage{subcaption}
\usepackage{epstopdf}

\newtheorem{remark}{Remark}
\newtheorem{lemma}{Lemma}

\title{Multi-agent Aerial Monitoring of Moving Convoys using Elliptical Orbits}

\author{Aseem V. Borkar$^{1}$
					and Girish Chowdhary$^{2}$
\thanks{$^{1}$Aseem V. Borkar is a Postdoctoral Research Associate at the Coordinated Science Laboratory, University of Illinois at Urbana Champaign, USA. {\tt\small avborkar@illinois.edu}}
\thanks{$^{2}$Girish Chowdhary is an Associate Professor of Agricultural and Biological Engineering and Computer Science, University of Illinois Urbana-Champaign, USA. {\tt\small girishc@illinois.edu}}
}

\begin{document}

\maketitle

\thispagestyle{empty}
\pagestyle{empty}

\begin{abstract}
We propose a novel scheme for surveillance of a dynamic ground convoy moving along a non-linear trajectory, by  aerial agents that maintain a uniformly spaced formation on a time-varying elliptical orbit encompassing the convoy. Elliptical orbits are used as they are more economical than circular orbits for circumnavigating the group of targets in the moving convoy. The proposed scheme includes an algorithm for computing feasible elliptical orbits, a vector guidance law for agent motion along the desired orbit, and a cooperative strategy to control the speeds of the aerial agents in order to quickly achieve and maintain the  desired formation. 
It achieves mission objectives while accounting for  linear and angular speed constraints on the aerial agents. The scheme is validated  through simulations and actual experiments with  a convoy of ground robots and a team of quadrotors as the aerial agents, in a motion capture environment.  
\end{abstract}

\section{Introduction}
\label{sec_intro}

We consider the problem of autonomously monitoring a  convoy with a group of aerial agents. A convoy is a group of ground vehicles trailing one behind the other on a common path. The problem of tracking ground targets using aerial agents has been well studied for both single and multiple Unmanned Aerial Vehicles (UAVs). Most solutions proposed earlier focus on circumnavigating the targets using  circular orbits. However, for circumnavigating a moving convoy,  elliptical orbits are often more economical to traverse as compared to circular orbits. The convoy monitoring problem has generally been studied for straight line paths or for paths whose curvature is known a priori. In this work, we propose a cooperative strategy for multiple aerial agents to traverse a time varying elliptical orbit around a moving convoy while maintaining an equi-distributed formation, without any knowledge of the convoy path. 

In \cite{borkarborkarcdc} and \cite{borkarborkarejcon}, the authors proposed an algorithm to compute a time varying elliptical orbit encompassing the convoy and a vector-field guidance law to drive a single aerial agent, modeled using unicycle kinematics, to track this orbit. For a multi-UAV mission, the objective is to achieve and maintain a multi-agent formation on an elliptical orbit that equally divides the  elliptical  orbit in terms of the parametric length of the ellipse. The original contributions of this work are:\\
1. Improvements to the vector field guidance law for better tracking of  the desired elliptical orbit.\\
2. An updated algorithm for computing feasible elliptical orbits to encompass the target convoy, while accounting for the motion constraints of the aerial agents.\\
3. A strategy to control the linear speeds of the aerial agents to achieve uniform spacing on the elliptical orbit.\\
4. A cooperation scheme for fast convergence to the formation having equal parametric spacing on the elliptical orbit.

Our work is applicable to a large class of UAVs such as autonomous fixed wing aircraft, helicopters and quadrotors. This is because we approximate the aerial agents as unicycle kinematic agents with constraints on  minimum and maximum linear speed and maximum angular speed.  We validate our results with MATLAB simulations for unicycle agents, and for a quadrotor model using a ROS-Gazebo based software in the loop  (SITL) simulator. We have also implemented the proposed strategy with a team of quadrotors and a convoy of ground robots in a motion capture environment. 

The paper is organized as follows: Section \ref{sec_lit_review} reviews some related work. Section \ref{sec_problem} describes the problem setting,  assumptions and notation. Section \ref{sec_preliminatries} discusses some preliminary concepts that are used for developing the proposed strategy. Section \ref{sec_guidance_law} discusses the vector field guidance law used to drive the aerial agents to the elliptical orbit. In Section \ref{sec_linear_speed}, the linear speed profile of the aerial agents is designed to maintain the desired formation. Section \ref{sec_orbit_computation} discusses the scheme for calculating the time varying elliptical orbit around the convoy. In Section \ref{sec_cooperation}, the main cooperation algorithm for attaining the desired aerial formation is discussed. Section \ref{sec_sim_expt} validates the proposed strategy by simulations and actual experiments.


\section{Related Work}
\label{sec_lit_review}
This section is a short overview of related literature, see \cite{targetsurvey}  for an extended survey. Our proposed scheme focuses on  monitoring a  ground convoy by aerially circumnavigating it with UAVs using a vector field based guidance law to  track elliptical orbits encompassing the ground convoy. In literature such strategies are generally used for tracking circular orbits for UAV path-planning \cite{beard_journal} and atmospheric sensing \cite{vfield_cylinder} applications.  They have also been developed in \cite{vfieldratnoo} for tracking circular orbits around stationary and moving targets while  respecting curvature demands of the underlying vector field. For multiple UAVs monitoring a single  target \cite{frew_circle_standoff} or a closely moving group of targets \cite{tsourdos_journal} while flying on a circular orbit,  the UAVs are treated as unicycle kinematic agents and the phase separation is  maintained by controlling the linear speed of the UAVs. This  is close in spirit to our linear speed control strategy for maintaining a formation on the time varying elliptical orbit. 

The vector field guidance law in \cite{frew_circle_standoff} is extended to track a racetrack like orbit in \cite{frew_racetrack}, where the UAV alternates between tracking a  circular segment and a straight line. Frew \cite{frew_ellipse} extends it to tracking an  elliptical orbit  whose parameters depend on estimation uncertainties of the target states. In both, the UAVs track  a target moving in a straight line. In contrast, our  strategy guides the aerial agents to track  a flexible ellipse that adapts its shape and orientation to the convoy trajectory.

Another approach for aerial monitoring of  ground targets is the cooperative cyclic pursuit, where multiple aerial agents that follow unicycle kinematics  cooperatively converge  to a circular orbit with an equispaced formation around a ground target, both stationary \cite{galloway_cyc_pursuit} and moving \cite{ma_moving_cyc_pursuit}. In \cite{ma_moving_var_rad}, such a cooperative scheme is proposed for achieving a circular formation on orbits having varying radii around the target. Our scheme generalizes to ellipses and circular orbits of varying sizes and gives a more robust orbit tracking as a result of its vector field guidance law.

Other approaches to formation control include \cite{zhang_formation} and \cite{leonard_formation}, where the agents switch between revolutions on a circular orbit  around slow moving targets and fixed formation paths to  move with faster moving targets, e.g., to track a point moving on a piece-wise linear path. 
 In \cite{zhousatya}, linear and circular paths are considered for the target and the emphasis is on  analyzing the error caused by wind.  In \cite{Kap}, the target is assumed to follow a smooth trajectory specified a priori as the level curve of a known function. In \cite{muslimovadaptive}, a backstepping-based controller is designed to follow a vector field to achieve a multiUAV formation on a circular orbit around a target while respecting the input constraints of the UAVs. Most of these approaches are designed towards tracking circular orbits. For a convoy of ground targets, an elliptical orbit can be more economical in terms of path length, see \cite{borkarborkarcdc}, \cite{borkarborkarejcon} for the single agent case. A consensus based strategy for elliptical formation around a ground target is proposed in \cite{ma2019ellipse}. Here the agent models considered are single and double integrator. Our strategy, for a unicycle type aerial agent with input constraints, is applicable for a large class of UAV platforms.

\section{Problem Description and Notation}
\label{sec_problem}
We assume a target ground convoy of $N_T$ vehicles trailing one behind the other on some unknown common path. Each vehicle moves with a maximum speed of $V_{T_{max}}$. We number the vehicles as $1,2,\dots,N_T$ along the direction of travel of the convoy with vehicle $N_T$ being the lead vehicle. Our objective is to design a cooperation strategy for a team of $N_A$ aerial agents for monitoring the moving convoy. The aerial agents are modeled using unicycle kinematics flying at a commanded altitude as follows:
\begin{align}
\dot{x}_{Ai}(t)=V_{Ai}\cos(\psi_{Ai}(t)) &,\
\dot{y}_{Ai}(t)=V_{Ai}\sin(\psi_{Ai}(t)),\nonumber\\
\dot{\psi}_{Ai}(t)&=\omega_{Ai}(t),
\label{eqn_unicycle}
\end{align}
where for agent $i\in\{1,2,...,N_A\}$, $(x_{Ai}(t),y_{Ai}(t))$ are the position coordinates, $\psi_{Ai}(t)$ is the heading angle, $V_{Ai}\in[V_{A_{min}},\ V_{A_{max}}]$ is the commanded linear speed, and $\omega_{Ai}(t)$ is the commanded angular velocity of  the monitoring agent satisfying $\vert\omega_{Ai}(t)\vert \leq \omega_{max} $. We make the following assumptions:\\
1.  The upper bound on convoy speed $V_{T_{max}}$ and  the convoy vehicle positions $(x_{Ti}(t),y_{Ti}(t))$  for all $i \in \{1,\dots,N_T\}$ are known to the aerial agents, e.g., through communication for a cooperating convoy or through on board sensors.\\
2. The maximum speed of the ground convoy and the linear speed bounds of the aerial agents are related as follows:
\begin{align}
0\leq V_{T_{max}}<<V_{A_{min}}< V_{A_{max}}-2V_{T_{max}}.
\label{eqn_speed_assumption}
\end{align} 
That is, the aerial agents are assumed to have a sufficient speed advantage over the ground convoy, which is reasonable if they have to circumnavigate the moving convoy. \\
3. The aerial agents have a sufficiently large communication range for effective cooperation.
\\
4. The aerial agent altitudes are high enough to ignore terrain.

We denote  the rotation matrix from the right handed global reference frame to a tilted frame with tilt angle $\theta(t)$  as
\begin{align}R_{\theta}(t)=\left[\begin{matrix}
\cos(\theta(t)) &  \sin(\theta(t))\\
  -\sin(\theta(t)) & \cos(\theta(t))
\end{matrix}\right].
\label{eqn_rot_mat}
\end{align}

\section{Preliminaries}
\label{sec_preliminatries}
We now discuss some useful properties of elliptical orbits, used later for developing the guidance law and the cooperation strategy between aerial agents for achieving the desired formation. The parametric equation of the ellipse is:
\begin{align}
x(s)=a\cos(s),\ y(s)=b\sin(s).
\label{eqn_ellipse_param}
\end{align} 
Here $s$ is a parameter taking values in $\left[0,\ 2\pi\right)$, $a$ is the semi-major axis length and $b$ is the semi-minor axis length, with $a\geq b>0$.
The equation of the ellipse then is $\nicefrac{x(s)^2}{a^2}+\nicefrac{y(s)^2}{b^2}=1$.

The first and second time-derivatives of  \eqref{eqn_ellipse_param} are \\
$\dot{x}(s)=-a\sin(s)\dot{s},\  \dot{y}(s)=b\cos(s)\dot{s},$\\
$\ddot{x}(s)=-a\cos(s)\dot{s}^2-a\sin(s)\ddot{s},$\\$ \ddot{y}(s)=-b\sin(s)\dot{s}^2+b\cos(s)\ddot{s}$.
 
  If a point moves on the ellipse with a parametric speed $\dot{s}>0$, then, the actual speed of the point is given by 
\begin{align}
V(s)=\sqrt{\dot{x}(s)^2+\dot{y}(s)^2}= \sqrt{ a^2\sin^2(s)+b^2\cos^2(s)}\dot{s}.
\label{eqn_actual_speed} 
 \end{align}
 
Similarly the curvature and radius of curvature of the ellipse at parameter value $s$ are given by
\begin{align}
\kappa(s)&=\tfrac{| \dot{x}(s)\ddot{y}(s)- \dot{y}(s)\ddot{x}(s) |}{(\dot{x}(s)^2+\dot{y}(s)^2)^{\frac{3}{2}}}=\tfrac{ab}{( a^2\sin^2(s)+b^2\cos^2(s))^{\frac{3}{2}}},
\label{eqn_curvature}\\
R(s)&=\kappa^{-1}(s)=\tfrac{( a^2\sin^2(s)+b^2\cos^2(s))^{\frac{3}{2}}}{ab}.
\label{eqn_r_curvature}
\end{align}  
 
 Define $G(s)=\sqrt{ a^2\sin^2(s)+b^2\cos^2(s)}$. Successively differentiating $G(s)$ with respect to $s$, we have
 \begin{align}
 G^{(1)}(s) &= \nicefrac{(a^2-b^2)\sin(2s)}{2\sqrt{ a^2\sin^2(s)+b^2\cos^2(s)}},
\label{eqn_dGds} \\
 G^{(2)}(s)&= \nicefrac{(a^2-b^2)\cos(2s)}{G(s)}   -   \nicefrac{(a^2-b^2)^2\sin^2(2s)}{4G(s)^3}.
\label{eqn_d2Gds2} 
    \end{align}
By the first order necessary conditions for  maxima or minima of $G$, setting $ G^{(1)}(s)=0$ gives the following cases:
 
\noindent {\it Case 1}: If  $a=b$, then the ellipse is a circle with   $ G^{(1)}(s)=0$, for all $s\in[0,\ 2\pi)$, i.e., $G(s)=K$ where $K$ is a constant.
 
   \noindent {\it Case 2}: If $a>b$, then $\sin(2s)=0$, so $s=\nicefrac{(2n+1)\pi}{2},\ n\pi$ for $n\in \{0,1\}$, as $s\in [0,\ 2\pi)$. By the second order sufficiency condition for maxima or minima of $G$, from \eqref{eqn_d2Gds2}, 
\begin{align*} 
 G^{(2)}(n\pi)=\frac{a^2-b^2}{b}>0,\  
 G^{(2)}\left(\tfrac{(2n+1)\pi}{2}\right)=\frac{b^2-a^2}{a}<0. 
\end{align*}  
Thus $s=n\pi$ minimizes and $s=\nicefrac{(2n+1)\pi}{2}$  maximises $G(s)$.    
   
\begin{figure}[!h]
\centering
\includegraphics[width=0.9\linewidth]{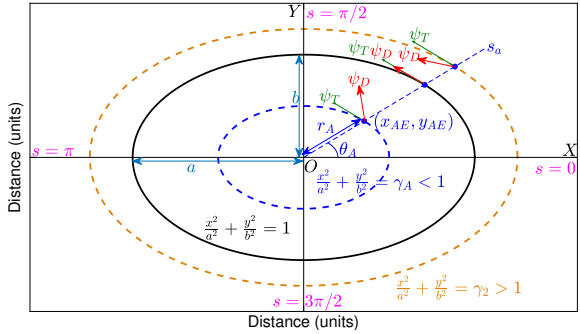}\\
\caption{Geometric representation of  $(x_{AE},y_{AE})$, $(r_A,\theta_A)$ and $(s_A,\gamma_A)$. The $s$ values listed in Table \ref{tab_extrema} are shown in magenta. The desired orbit is plotted in solid black. The commanded aerial agent heading  and local tangential heading shown in red and green respectively.}
\label{fig_ellipse_param}
\end{figure}   
   
Consider a point moving with constant parametric speed $\dot{s}=s_v$  on the ellipse. From \eqref{eqn_actual_speed}, \eqref{eqn_curvature} and \eqref{eqn_r_curvature}, $V(s)=G(s)s_v$, $\kappa(s)=\nicefrac{ab}{G(s)^3}$ and $R(s)=\nicefrac{G(s)^3}{ab}$ respectively, which implies that $s=n\pi$ minimizes $V(s)$ and $R(s)$ and maximizes $\kappa(s)$, and vice versa for   $\nicefrac{(2n+1)\pi}{2}$. Table \ref{tab_extrema} summarizes the maximum and minimum values of $V(s)$, $R(s)$ and $\kappa(s)$.
   \begin{table}[h!]
\centering
\caption{Extrema values on an ellipse for a point moving with constant parametric speed $\dot{s}=s_v$ and $n\in\{0,1\}$}
\label{tab_algo_ip}
\resizebox{8.5cm}{!}{
\begin{tabular}{|c|c|c|c|}
\hline
$s $                   & $V(s)$            & $R(s)$                  & $\kappa(s)$                  \\ \hline
$n\pi$                & $V_{min}=b s_v $ & $R_{min}=\nicefrac{b^2}{a}$ & $\kappa_{max}=\nicefrac{a}{b^2}$ \\ \hline
$\nicefrac{(2n+1)\pi}{2}$ & $V_{max}=a s_v $ & $R_{max}=\nicefrac{a^2}{b}$ & $\kappa_{min}=\nicefrac{b}{a^2}$ \\ \hline
\end{tabular}}
\label{tab_extrema}
\end{table}
\subsection*{Parameter computation for the aerial agents}
If the aerial agents start with equi-parametric separations  on the ellipse, then they can maintain it by moving at a constant parametric speed $\dot{s}=s_v$. But it is not necessary that they  start  on the desired elliptical orbit. For an agent at position $(x_{AE},y_{AE})$ expressed in the orbit-centered coordinate frame, we need  to calculate the parameter value $s_A$ for the agent. Given a desired orbit centered at origin $O$ with axis constants $a$ and $b$, every point in space lies on an ellipse of the form $\frac{x^2}{a^2}+ \frac{y^2}{b^2} = \gamma$ concentric with the desired orbit, where $\gamma<1$ for points inside the desired orbit, $\gamma=1$ for points on it, and $\gamma>1$ for points outside, as shown in Fig.\ \ref{fig_ellipse_param}. Then the parametric equation for these concentric ellipses is given by
\begin{align}
x(\gamma,s)=a\sqrt{\gamma}\cos(s),\ y(\gamma,s)=b\sqrt{\gamma}\sin(s).
\label{eqn_ellipse_param_gamma}
\end{align}

Thus the coordinates of the point $(x_{AE},y_{AE})$ can be represented by \eqref{eqn_ellipse_param_gamma} for some value of $s_A$ and $\gamma_A$.  If the corresponding polar coordinates of  $(x_{AE},y_{AE})$ in the orbit-centric reference frame are  $(r_{A},\theta_{A})$, then 
\begin{align}
\theta_{A}=\arctan2(y_{AE},x_{AE}),
\label{eqn_theta_a}
\end{align}
and  $(x_A,y_A)=(r_A\cos(\theta_A),r_A\sin(\theta_A))$. Equating these with $(x(\gamma_A,s_A),y(\gamma_A,s_A))$ in \eqref{eqn_ellipse_param_gamma}, we have  $a\sqrt{\gamma_A}\cos(s_A)=r_A\cos(\theta_A)$ and $b\sqrt{\gamma_A}\sin(s_A)=r_A\sin(\theta_A)$. Hence 
 $\frac{b}{a}\tan(s_A)=\tan(\theta_A)$. Thus $s_A$ is computed as 
 \begin{align}
s_A=d_c\arctan2(a\sin(\theta_A),b\cos(\theta_A)), 
\label{eqn_s_theta}
 \end{align}
 where $d_c$ is a direction selection input chosen as
\begin{align}
d_c=\begin{cases} -1, \ \ \ \mbox{for clockwise motion on orbit}\\
1, \ \ \  \mbox{for counter-clockwise motion on orbit}.
\end{cases}
\label{eqn_dc}
\end{align}
Hence given agent's position $(x_{AE},y_{AE})$ and desired orbit constants $a, b$, we can compute the parameter $s_A$  on the concentric ellipse given by \eqref{eqn_ellipse_param_gamma}, using \eqref{eqn_theta_a}  and \eqref{eqn_s_theta}. The value of $\gamma_A$ is given by 
\begin{align}
\gamma_A=\frac{x_{AE}^2}{a^2}+ \frac{y_{AE}^2}{b^2} .
 \label{eqn_gamma}
\end{align}
From \eqref{eqn_s_theta}, it can be shown that $s_A=d_c\theta_A$, when $\theta_A=n\pi $
 and $\nicefrac{(2n+1)\pi}{2}$ for $n \in \{0,1\}$, as illustrated in Fig.\ \ref{fig_ellipse_param}. 
\section{Vector-field guidance law}
\label{sec_guidance_law}
The vector field guidance law computes a desired heading command that guides the aerial agent towards the desired elliptical orbit described by \eqref{eqn_ellipse_param}. Our guidance law  is an update of that in \cite{borkarborkarejcon}, with a modification for improved tracking performance. If $(x_{AE},y_{AE})$ are the agent position coordinates in the coordinate frame centered and aligned with  the desired elliptical orbit (see Fig.\ \ref{fig_ellipse_param}), then the commanded heading $\psi_D$ of the aerial agent has two components remove space: $\psi_T$ is the tangential direction to the concentric ellipse described by \eqref{eqn_ellipse_param_gamma} at position $(x_{AE},y_{AE})$  as in Fig.\ \ref{fig_ellipse_param}, and $\psi_O$ is an offset term responsible for generating the  heading command $\psi_D$ to reach the desired orbit. 
These are computed as
\begin{align}
\psi_T&= \arctan2\left(d_c b^2x_{AE},-d_c a^2y_{AE}\right), \nonumber\\
\psi_{O}  &=d_c\arctan\left(k_\gamma\kappa(s_A)(\gamma_A -1)\right),\nonumber\\
\psi_D&=\psi_T+\psi_O.
\label{eqn_psi_D}
\end{align}
     $d_c$ is chosen corresponding to the direction of motion along the elliptical orbit, according to \eqref{eqn_dc}. $k_\gamma>0$ is a gain used as tuning parameter. $s_A,\gamma_A$ and $\kappa(s_A)$  are computed using \eqref{eqn_s_theta}, \eqref{eqn_gamma} and \eqref{eqn_curvature} respectively.  
As shown in Fig.\ \ref{fig_ellipse_param}, $\psi_O$ adds an appropriate offset to $\psi_T$, such that the resulting heading command $\psi_D$ drives the agent towards the desired elliptical orbit.
The aerial agents track  $\psi_D $ by commanding the angular velocity of the aerial agent as follows:
\begin{align}
\omega_A(t) \  = k_\psi (\psi_D(t)-\psi_{AE}),
\label{eqn_omg_a}
\end{align}
where $k_\psi$ is a control gain, $\psi_{AE}=\psi_{A}-\theta_E$ is the heading angle of the aerial agent expressed relative to the orbit centric reference frame, and  $\theta_E$ is the tilt angle of the ellipse, relative to the  global reference frame. The angular velocity  of the aerial agent given by \eqref{eqn_omg_a} is constrained as  $\vert\omega_{Ai}(t)\vert \leq \omega_{max} $. For proof of convergence of the guidance law  to the desired elliptical orbit, see  \cite{borkarborkarejcon}, as  the same proof applies here.

\begin{figure}[h]
\centering
\includegraphics[width=1\linewidth]{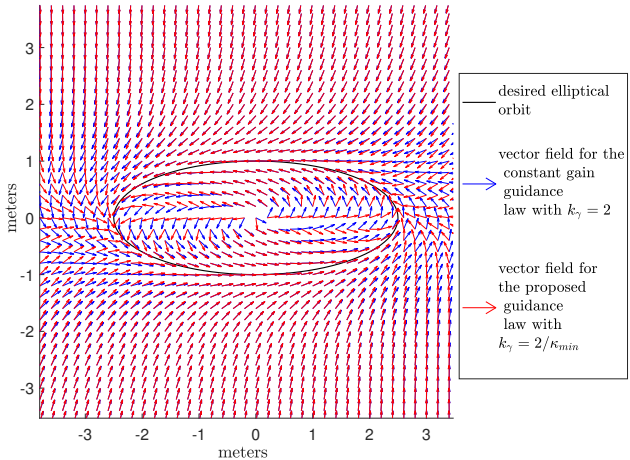}\\
\caption{Comparison between the vector fields for constant gain guidance law in \cite{borkarborkarejcon} and the proposed guidance law given by \eqref{eqn_psi_D}}
\label{fig_vfield_compare}
\end{figure}

In \eqref{eqn_psi_D}, we improve the guidance law in \cite{borkarborkarejcon} by introducing the curvature term $\kappa(s_A)$ in the $\psi_O$ to increase the  sensitivity of the guidance law in the  high curvature regions of the elliptical orbit. This yields in better orbit tracking than the guidance law  in \cite{borkarborkarejcon}.
We demonstrate the improvement in the tracking performance of this updated  guidance law simulating  two unicycle agents moving with  a constant speed  $V_A=0.4$ m/sec, $\vert\omega_A\vert\leq 1.5$ rad/sec, $k_\psi=1$ and $d_c=1$.
Both track an elliptical orbit centered around the origin with $a=2.5$ m, $b=1$ m. From Table \ref{tab_extrema}, the minimum curvature  is $\kappa_{min}= 0.16 $ m$^{-1}$. For a fair comparison, the agent $1$ uses the guidance law in \cite{borkarborkarejcon} with $k_\gamma=2$, and   the agent $2$ uses the guidance law given by \eqref{eqn_psi_D}, with the gain  $k_\gamma=2/\kappa_{min}$. 
\begin{figure}[h]
\centering
\includegraphics[width=1\linewidth]{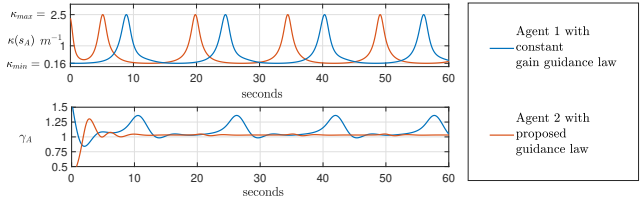}\\
\caption{Comparison between the constant gain guidance law in \cite{borkarborkarejcon} and the updated guidance law in \eqref{eqn_Ds}. For ideal tracking $\gamma_A=1$.}
\label{fig_guidance_comparison}
\end{figure} 
From the plots in Fig. \ref{fig_guidance_comparison}, using $\gamma_A$  (given by \eqref{eqn_gamma}) as the  metric, we see that the performance of agent $2$ is better than agent $1$  particularly in tracking the higher curvature regions of the orbit.
 We show a  comparison of the  the underlying vector fields  in Fig. \ref{fig_vfield_compare} for guidance laws used for the above simulated agents.

We can see that in the low curvature regions the vector fields are nearly identical, thus for initial conditions corresponding to trajectories reaching the orbit in the low curvature regions the convergence rate would be similar. In the high curvature regions the updated guidance law will result in faster convergence to the desired orbit due to the tighter vector-field.  This gives the proposed guidance better tracking performance in the high curvature regions once the orbit is reached.

\section{Linear speed profile for aerial agents}
\label{sec_linear_speed}
In \cite{borkarborkarcdc,borkarborkarejcon} the single aerial agent monitoring the convoy was commanded to fly with a constant linear speed, and the angular speed was controlled so as to guide the agent to the desired elliptical orbit. Switching to the multi-agent framework, the objective now is to obtain a parametrically  equispaced formation of $N_A$ aerial  agents on the ellipse. That is, the parametric separation between any pair of consecutive aerial agents on the ellipse is  $\delta_s=\nicefrac{2\pi}{N_A}$. This is done by controlling the linear speeds of the aerial agents. 

The objective of the linear speed control  is to achieve and maintain an aerial agent formation that divides the parametric length of the ellipse ($2\pi$) into $N_A$  equal parts. To do this the commanded linear speed for each agent comprises of three components:\\ 1) Nominal speed profile $V_{Ei}$\\ 2) Correction component $V_{Ci}$\\ 3) Motion compensation component   $\bar{V}_{T} $.

Agents moving with equal constant parametric speed $\dot{s}_A$ on  an ellipse maintain their initial  parametric separation on the ellipse. Thus we first compute a nominal linear speed profile for the agent corresponding  to the  constant parametric speed $\dot{s}_A=s_v$ for motion on the stationary elliptical orbit described by \eqref{eqn_ellipse_param_gamma}. From Table \ref{tab_extrema},  the maximum and minimum speeds on the  ellipse  for such a motion are $V_{max}=a\sqrt{\gamma_A}s_v$ and  $V_{min}=b\sqrt{\gamma_A}s_v$. Suppose we want to constrain $V_{min}$, $V_{max}$  to a given interval $[V_{E_{min}},\  V_{E_{max}}]$, then in order to symmetrically utilize the interval, we select $s_v$ such that the average speed $\nicefrac{(V_{min}+V_{max)}}{2}$ resulting from constant parametric rate $\dot{s}=s_v$ equals $\nicefrac{(V_{E_{min}}+V_{E_{max}})}{2}$, leading to 
\begin{align}
 s_v=\frac{V_{E_{min}}+V_{E_{max}}}{(a+b)\sqrt{\gamma_A}}.
\label{eqn_sv}
\end{align}
 
 To ensure that the motion due to this parametric rate does not violate the given constraint, we need $[V_{min}, V_{max}]\subseteq  [V_{E_{min}},\  V_{E_{max}}]$. This leads to:
\begin{align}
 \frac{V_{E_{min}}}{b\sqrt{\gamma_A}} \leq s_v &\mbox{ and } s_v \leq \frac{V_{E_{max}}}{a\sqrt{\gamma_A}} 
 \Rightarrow \frac{V_{E_{min}}}{V_{E_{max}}}\leq  \frac{b}{a}.
 \label{eqn_ab_Ve_constraint}
\end{align}
We select $a,b$ in Section \ref{sec_orbit_computation} such that   \eqref{eqn_ab_Ve_constraint} is satisfied. The nominal  linear speed corresponding to the constant parametric rate $s_v$  for an aerial agent $i$ at position coordinates $(x_{AEi},y_{AEi})$, moving on a stationary ellipse described by \eqref{eqn_ellipse_param_gamma}, is now computed using \eqref{eqn_actual_speed} as
\begin{align}
V_{Ei}(s_{Ai},\gamma_{Ai})= \sqrt{\gamma_{Ai}( a^2\sin^2(s_{Ai})+b^2\cos^2(s_{Ai}))}s_v,
\label{eqn_VE}
\end{align}
where  $s_{Ai}$ is calculated using \eqref{eqn_theta_a} and \eqref{eqn_s_theta}, and $\gamma_{Ai}$ and $\dot{s}_{Ai}=s_{vi}$ computed  using \eqref{eqn_gamma}  and  \eqref{eqn_sv} respectively. Then from \eqref{eqn_sv} and assuming \eqref{eqn_ab_Ve_constraint} is satisfied,  $V_{Ei}\in [V_{E_{min}},\  V_{E_{max}}]$.

To drive the parametric spacing between neighboring agents to the desired separation of $\nicefrac{2\pi}{N_A}$ on the ellipse, we add a control component $V_{Ci}$ to the nominal speed $V_{Ei}$ of the agents. $V_{Ci}$ is computed based on cooperation with a neighboring agent as explained in the Section \ref{sec_cooperation}. The resulting linear speed magnitude command then is 
\begin{align}
\tilde{V}_{Ai}=V_{Ei}(s_{Ai},\gamma_{Ai})+V_{Ci}.
\label{eqn_VA_ellipse_centric}
\end{align}

For monitoring a moving convoy, the elliptical orbit is not stationary. We have designed the speed profile $\tilde{V}_{Ai}$ in the orbit centric frame, and to compensate for orbit motion we add the  the orbit center velocity to $\tilde{V}_{Ai}$ to compute the linear speed command relative to the global reference frame. We assume that the center of the ellipse moves with a velocity vector $(\bar{V}_{Tx},\ \bar{V}_{Ty})$, with a magnitude bounded above by $V_{T_{max}}$. In Section \ref{sec_orbit_computation} we propose orbit computation strategy to guarantee this.
%
%
%
%
Since the aerial agents  follow unicycle kinematics, their linear velocity components along the axes of the global reference frame cannot be commanded separately due to the non-holonomic constraint on the unicycle model. Thus  an approximate compensation is used where the  magnitude of the resultant vector is commanded as the linear speed of the agent $i$ given by
\begin{align}
V_{Ai}=\sqrt{(\tilde{V}_{Ai}\cos(\psi_{Ai})+\bar{V}_{Tx})^2+(\tilde{V}_{Ai}\sin(\psi_{Ai})+\bar{V}_{Ty})^2}.
\label{eqn_linear_speed_cmd}
\end{align}

Since the motion compensation component can at most  increase or decrease the agent speed by $V_{T_{max}}$, the range of speeds available to design $\tilde{V}_{Ai}$ is $[V_{R_{min}},\   V_{R_{max}}]$, where $V_{R_{max}}=V_{A_{max}}-V_{T_{max}}$ and $V_{R_{min}} =V_{A_{min}}+V_{T_{max}}$. This interval is not empty due to assumption \eqref{eqn_speed_assumption}.  


In order to allow for sufficient band of speeds above and below the nominal linear speed profile given by \eqref{eqn_VE} for the cooperative component $V_{Ci}$, we  select a value $\delta\in (0, 1]$, and   bounds $V_{E_{min}}$ and $V_{E_{max}}$  on the nominal speed profile \eqref{eqn_VE} such that:
\begin{align*}
\nicefrac{(V_{E_{min}}+V_{E_{max}})}{2}&=\nicefrac{(V_{R_{max}} +V_{R_{min}})}{2}, \nonumber \\
V_{E_{max}}-V_{E_{min}}&=\delta(V_{R_{max}}-V_{R_{min}}).
\end{align*}
Solving this system of linear equations yields
\begin{align}
V_{E_{min}}=(1-\delta)\nicefrac{V_{R_{max}}}{2}+(1+\delta)\nicefrac{V_{R_{min}}}{2},\\
V_{E_{max}}=(1+\delta)\nicefrac{V_{R_{max}}}{2}+(1-\delta)\nicefrac{V_{R_{min}}}{2}.
\end{align}
The intervals $[V_{E_{max}},\ V_{R_{max}}]$ and $[ V_{R_{min}},\ V_{E_{min}}]$ allow room for the action of $V_{Ci}$. 
Finally, the total linear speed command $V_{Ai}$ given by \eqref{eqn_linear_speed_cmd} is saturated so as to maintain $V_{Ai}\in[V_{A_{min}},\ V_{A_{max}}]$, if the contribution of $V_{Ci}$ is large.  
 
\section{Elliptical orbit computation}
\label{sec_orbit_computation}
As discussed in Section \ref{sec_linear_speed} we need an algorithm for computing the elliptical orbit for encompassing the ground convoy, which guarantees that  orbit center velocity is bounded above by $V_{T_{max}}$. The orbit computation scheme proposed in \cite{borkarborkarejcon} cannot be used directly as it  does not enforce this constraint. We thus propose an alternate scheme to  compute the desired orbit centered at the target centroid  $(\bar{x}_T,\bar{y}_T)$ (average position of the ground convoy vehicles). Since the speed of the ground convoy vehicles is bounded above by $V_{T_{max}}$, the magnitude of velocity of the target centroid is also bounded above by $V_{T_{max}}$. 
%
We choose a reference  line parallel to the line segment joining the first and last vehicle positions  in the convoy (i.e., convoy targets with indices $i=1$ and $i=N_T$), and passing through the target centroid  $(\bar{x}_T,\bar{y}_T)$, which is the mean position of the convoy vehicles.
 The angle made by the reference line with respect to the global reference frame is denoted by $\theta_E=\arctan2(y_{TN_T}-y_{T1},\ x_{TN_T}-x_{T1})$, which is the tilt angle of the elliptical orbit.
\begin{figure}[!h]
\centering
\includegraphics[width=.9\linewidth]{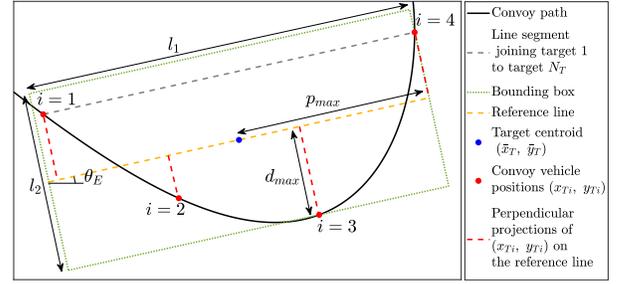}\\
\caption{Bounding box  for a target convoy with $N_T=4$ }
\label{fig_bbox}
\end{figure}

The perpendicular projections from  the convoy target positions to this reference line are computed. The length of the bounding box is selected as $l_1=2p_{max}$, where $p_{max}$ is the distance between the target centroid and the farthest projection point on the reference line. The breadth is selected as $l_2=2d_{max}$, where $d_{max}$ is the maximum perpendicular distance of the convoy  vehicle positions from the reference line. The bounding box is centered at the target centroid $(\bar{x}_T,\bar{y}_T)$. An example of this construction for a convoy of $N_T=4$ vehicles is shown in Fig.\ \ref{fig_bbox}. Since this strategy is implemented iteratively with $k$ being the iteration count, and a constant loop delay $\Delta t$, Algorithm \ref{algo_CCR} takes current convoy position values as inputs and computes the values of $l_1(k),\ l_2(k)$.
\begin{algorithm}[!h]
\caption{\textproc{Bounding$\_$box$\_$computation}}
\label{algo_CCR}{\small
\begin{algorithmic}[1]
\Statex {\bf Inputs: } $(x_{Ti}(k),y_{Ti}(k)) \forall i\in\{1,...,N_T \} $ ,\Statex \qquad \qquad     $k$, $(\bar{x}_T(k),\bar{y}_T(k)),\theta_E(k)$
\Statex {\bf Outputs: } $l_1(k)$, $l_2(k)$ 
\Statex {\bf Initialise: } $x_{min}\leftarrow 0,x_{max}\leftarrow0$, $d_{max}\leftarrow0$, $p_{max}\leftarrow0$
\For {$i\in \{1,...,N_T \}$}
\State $\left[\begin{matrix}x_r\\ y_r \end{matrix}\right]\leftarrow R_{\theta_E}(k)\left[\begin{matrix} (x_{Ti}(k)-\bar{x}_T(k) \\ (y_{Ti}(k)-\bar{y}_T(k) \end{matrix}\right]$
\If{$d_{max}\leq \vert y_r \vert $}
\State $d_{max}\leftarrow \vert y_r \vert$
\EndIf
\If{$x_{min}\geq x_r$}
\State $x_{min}\leftarrow x_r$
\EndIf
\If{$x_{max}\leq x_r$}
\State $x_{max}\leftarrow x_r$
\EndIf
\EndFor
\State $p_{max}\leftarrow \max(x_{max},\vert x_{min}\vert)$
\State $l_1(k)\leftarrow 2p_{max}$
\State $l_2(k)\leftarrow 2d_{max}$
\end{algorithmic}}
\end{algorithm}
To select the ellipse that contains this bounding box, we consider the minimum area ellipse circumscribing it according to the following Lemma.

\begin{lemma}(\cite{borkarborkarejcon})
The area of an ellipse $\mathcal{E}:\nicefrac{x^2}{a^2}+\nicefrac{y^2}{b^2}=1$ circumscribing a rectangle of dimensions $l_1\times l_2$ with $l_1>l_2>0$ is minimized by $a=\nicefrac{l_1}{\sqrt{2}}$ and $b=\nicefrac{l_2}{\sqrt{2}}$.
\label{lem_min_A}
\end{lemma}
We now have the following constraints on $a$ and $b$:\\
1. To ensure that the elliptical orbit is feasible, its minimum radius of curvature given in Table \ref{tab_extrema} should exceed the worst case minimum turn radius of the aerial agents, i.e., \\
$R_{min}\geq \frac{V_{A_{max}}}{\omega_{max}}\Rightarrow b\geq \sqrt{a\frac{V_{A_{max}}}{\omega_{max}}}$.\\
2. Since $a\geq b$,
$a \geq \sqrt{a\frac{V_{A_{max}}}{\omega_{max}}}\Rightarrow a\geq \frac{V_{A_{max}}}{\omega_{max}}$.\\
3. The constraint on $a$ and $b$ given by \eqref{eqn_ab_Ve_constraint}.\\ Imposing  these constraints along with Lemma \ref{lem_min_A}, we select the major and minor axes of the elliptical orbit as:
 \begin{align}
a(k)&=\max\left(\tfrac{l_1(k)}{\sqrt{2}}, \tfrac{l_2(k)}{\sqrt{2}},\tfrac{V_{A_{max}}}{\omega_{max}}\right),\\
b(k)&=\max\left(\tfrac{l_1(k)}{\sqrt{2}}, \tfrac{a(k)V_{E_{min}}}{V_{E_{max}}},\sqrt{\tfrac{a(k)V_{A_{max}}}{\omega_{max}}}\right).
\label{eqn_b_Rmin_constraint}
\end{align}
The  target centroid velocity is computed by filtering the target centroid position using an exponential smoothing filter, and then computing the discrete derivative $(\bar{V}_{Tx},\ \bar{V}_{Ty})=(\nicefrac{\Delta \bar{x}_T}{\Delta t},\nicefrac{\Delta \bar{y}_T}{\Delta t})$. This is then used as compensation in \eqref{eqn_linear_speed_cmd}.

\section{Cooperation for equi-parametric formation}
\label{sec_cooperation}
 In cooperative approaches for target detection such as cyclic pursuit, there is a possibility that the final multi-agent formation attained at equilibrium has more than one agent at the same position on the orbit around the target (equi-parametric formations having separations $\Delta_s$ which are integer multiples of $\nicefrac{2\pi}{N_A}$). Such formations are generally initial condition dependent (as shown in Fig.\ 9 and 12 in \cite{padmaji2012time}) or can be caused by saturation of inputs.  Our cooperation scheme avoids such scenarios by first driving the agents to the desired equi-parametric equilibrium formation and then maintains this formation for the remaining mission. Also, the neighbors on the ellipse are not predefined and get assigned based on the order in which the agents reach the ellipse. This allows the agents to settle to the desired equi-parametric formation faster. 
 
 Each aerial agent indexed by $i\in\{1,\dots,N_A\}$ knows its own pose, i.e., $(x_{Ai},y_{Ai}, \psi_{Ai})$, using a  localisation source (GPS, motion capture, etc.), and computes its parameters $s_{Ai}$ and $\gamma_{Ai}$, using \eqref{eqn_theta_a},  \eqref{eqn_s_theta} and  \eqref{eqn_gamma}. In order to achieve an equi-parametric formation, each aerial agent repeatedly broadcasts a communication packet (using the \textproc{Broadcast$\_$packet} function in Algorithm \ref{algo_coop}) containing the following:
\begin{enumerate}
\item Its identification index $i \in \{1,\dots,N_A\}$. 
\item Agent parameter $s_{Ai}$. 
\item Ready flag $fl_{Ri}$.
\item Formation height flag $fl_{Hi}$.
\item Orbit flag $fl_{Oi}$.
\end{enumerate}
For each such packet received, the \textproc{Receive$\_$packet} function in Algorithm \ref{algo_coop} updates the array entries in $ s_{A_{arr}}, fl_{R_{arr}},fl_{H_{arr}}$ and $fl_{O_{arr}}$ at the index locations corresponding to the agent index $i$ received in the packet. 

Before attaining the desired formation, in order to ensure collision free motion, all aerial agents are initialized to fly at different altitudes to ensure sufficient vertical separation. The guidance law of Section \ref{sec_guidance_law} drives all aerial agents to reach and latch onto  the desired orbit, thus for this discussion we assume $\gamma_{Ai}\approx 1$ for all agents once they reach the desired orbit. The agents start cooperation only after reaching the desired orbit. This is indicated by setting the  flag $fl_{O_i}$ when  error  $\vert \gamma_{Ai}-1\vert$ is within some threshold $\gamma_{Th}$ (say, $\gamma_{Th}=0.1$).  Also, each agent $i$ identifies the next agent $i_N$ on the desired elliptical orbit along the direction of motion on ellipse, based on the parameter  values  received in $ s_{A_{arr}}$. This is done in algorithm \ref{algo_coop} by the \textproc{Find$\_$neighbour} function. If two agents $i$ and $j$ with $j>i$ have the same parameter value, the tie is broken by assigning the $i_N=j$ for agent $i$.

 To ensure that the agents first attain the equispaced formation,  agent 1 acts as a reference leader and moves with a linear speed computed using \eqref{eqn_VA_ellipse_centric} and \eqref{eqn_linear_speed_cmd} with $V_{Ci}=0$. Thus    from  \eqref{eqn_VA_ellipse_centric}, in the orbit centered reference frame, agent $1$ moves with linear speed  $\tilde{V}_{A1}=V_{E1}(s_{A1}) \in [V_{E_{min}},\ V_{E_{max}}] $   corresponding to $\dot{s}_{A1}=s_v$.
The remaining agents compute the parametric separation error $D_{si}$ (error between the actual separation of agent $i$ from $i_N$, and the desired separation $\Delta_s=\nicefrac{2\pi}{N_A}$) and the resulting correction component  $V_{Ci}$ as
 \begin{align}
 D_{si}&=s_{Ai_N}-s_{Ai}-\Delta_s,
 \label{eqn_Ds}\\
V_{Ci}&=\sqrt{\gamma_{Ai}( a^2\sin^2(s_{Ai})+b^2\cos^2(s_{Ai}))}k_s D_{si}.
\label{eqn_Vc}
\end{align} 

The function \textproc{Normalize} in Algorithm \ref{algo_coop} is used to normalize the parameter differences in the interval $[-\pi,\ \pi]$. Suppose $i_N=1$ for an agent $i'$, and the separation $s_{A1}-s_{Ai'}>\Delta_s$, then the commanded  $V_{Ci'}>0$, which implies that $\tilde{V}_{Ai'}=V_{Ei'}(s_{Ai'})+V_{Ci'}$ corresponds to $\dot{s}_{Ai'}>s_v$, and vice-versa. Because of the choice of $\delta<1$ in Section \ref{sec_linear_speed},  $\tilde{V}_{Ai'}$ can take values less than $V_{E_{min}}$ or above $V_{E_{max}}$. This guarantees that agent $i'$ achieves and maintains desired separation relative to agent $1$. Upon achieving the desired separation $V_{Ci'}\approx 0$, and a similar arguments follows for the agent with  $i_N=i'$ and so on. Thus  convergence of all agents to the desired separation relative to agent $1$, is guaranteed.

\begin{algorithm}[!h]
\caption{\textproc{Aerial$\_$agent$\_$cooperation}}
\label{algo_coop}{\small
\begin{algorithmic}[1]
\Statex {\bf Inputs: } $i$, $s_{Ai}$, $\gamma_{Ai}$, $D_{Th}$
\Statex{\bf Functions: } \textproc{Find$\_$neighbour}, \textproc{Broadcast$\_$packet}, \textproc{Receive$\_$packet}, \textproc{Normalize}
\Statex{\bf Initialization: }$\Delta_s\leftarrow  \nicefrac{2\pi}{N_A}$, $fl_{Ri}\leftarrow \algofalse$, $fl_{Hi}\leftarrow \algofalse $,  $fl_{Oi}\leftarrow \algofalse $ for all $i\in\{1,\dots,N_A\}$
\State $[ s_{A_{arr}}, fl_{R_{arr}},fl_{H_{arr}},fl_{O_{arr}}]\leftarrow  \textproc{Receive \_ packet}()$
\State $i_N\leftarrow \textproc{Find$\_$neighbour}( s_{A_{arr}})$
\State $s_{Ai_N}\leftarrow  s_{A_{arr}}[i_N]$, $fl_{Ri_N}\leftarrow fl_{R_{arr}}[i_N]$, \Statex $fl_{Hi_N}\leftarrow fl_{H_{arr}}[i_N]$, $fl_{Oi_N}\leftarrow fl_{O_{arr}}[i_N]$ 
\State $D_s\leftarrow \textproc{Normalize}(\textproc{Normalize}(s_{Ai_N}-s_{Ai})-\Delta_s)$
\State $V_{Ci}\leftarrow 0$
\If {$\vert \gamma_{Ai} -1\vert<\gamma_{Th}$}
\State $fl_{Oi}\leftarrow \algotrue$
\EndIf
\If{$fl_{Oi}=\algotrue$ \algoand $fl_{Oi_N}=\algotrue$} 

\If{$i_N = 1$ \algoand $\vert D_s\vert<D_{Th}$}
\State $fl_{Ri}\leftarrow \algotrue$
\ElsIf {$fl_{Ri_N}=\algotrue$ \algoand $\vert D_s\vert<D_{Th}$}
\State $fl_{Ri}\leftarrow \algotrue$
\EndIf
\If {$fl_{Hi_N}=\algotrue$ }
\State $fl_{Hi}\leftarrow \algotrue$
\EndIf
\If{$i=1$}
\State $V_{Ci}\leftarrow 0$
	\If{$fl_{Ri_N}=\algotrue$} 
	\State $V_{Ci}\leftarrow \sqrt{\gamma_{Ai}( a^2\sin^2(s_{Ai})+b^2\cos^2(s_{Ai}))}k_sD_s$
	\State $fl_{Hi}\leftarrow \algotrue$
	\EndIf
\Else 	\State $V_{Ci}\leftarrow \sqrt{\gamma_{Ai}( a^2\sin^2(s_{Ai})+b^2\cos^2(s_{Ai}))}k_sD_s$
\EndIf
\EndIf
\State \textproc{Broadcast$\_$packet}($i,s_{Ai},fl_{Ri},fl_{Hi}$)

%
%

\end{algorithmic}}
\end{algorithm} 
When the desired formation is reached, $V_{Ci}\approx 0$ and $\tilde{V}_{Ai}\approx V_{Ei}(s_{Ai})$ for all agents, corresponding  $\dot{s}_{Ai}\approx s_v.$ Thus the inter-agent parametric separations are maintained. Each agent raises its ready flag $fl_{Ri}$ when $D_s$ is small (within a threshold of, say, $D_{Th}=0.1$), starting from the agent $i'$. Since the ellipse is a closed curve, once the remaining agents achieve the desired relative separations with respect to agent 1, the parametric separation between agent 1 and its neighbor $i_{N}$ is approximately $\Delta_s=\nicefrac{2\pi}{N_A}$. Thus in algorithm \ref{algo_coop}, when all the remaining agents have reached their formation positions relative to agent 1 and have set their ready flag $fl_{Ri}$, agent $1$ also sets its $fl_{Ri}$  and $fl_{Hi}$ flags and applies the control command given by \eqref{eqn_Vc}. The $fl_{Hi}$ flag is set throughout the formation to indicate that all agents should attain the same mission altitude for monitoring the convoy. As a result there is no longer a leader in the formation and all agent  maintain the formation about the stable equilibrium  of  $\tilde{V}_{Ai}=V_{Ei}(s_{Ai})$ using control $V_{Ci}$. 

  Algorithm \ref{algo_coop} gives the pseudo-code to implement the proposed cooperation strategy when a communication packet is received.

\section{Simulations and Experiments}
\label{sec_sim_expt}
\begin{table*}[!h]

\centering
\caption{Simulation and experiment cases and their parameters (Linear speeds in m/sec, angular speeds in rad/sec)}
\resizebox{18cm}{!}{
\large
\begin{tabular}{|l|l|l|l|l|l|l|l|l|l|l|l|l|l|}
\hline
                & $N_T$ & Convoy path     & Convoy speed profile & $V_{T_{max}}$ & $N_A$ & $V_{A_{min}}$ & $V_{A_{max}}$ & $\omega_{max}$ & $d_c$ & $k_s$ & $k_\psi$ & $k_\gamma$ & $\delta$ \\ \hline
MATLAB Simulation 1    & 6     & Curve           & Stationary           & 0             & 5     & 0.4           & 1.2           & 1.5            & 1     & 0.5   & 1.5      & 20         & 0.8      \\ \hline
MATLAB Simulation 2    & 5     & Lissajous curve & Time varying speed   & 0.2           & 6     & 0.3           & 1             & 1              & -1    & 0.5   & 1        & 20         & 0.8      \\ \hline
MATLAB Simulation 3    & 6     & Way-points & Constant speed       & 0.1           & 7     & 0.25          & 0.8           & 1              & 1     & 0.7   & 1        & 20         & 0.8      \\ \hline
SITL Simulation & 5     & Road network    & Constant speed       & 0.15          & 4     & 0.3           & 0.8           & 1              & 1     & 0.5   & 2        & 20         & 0.8      \\ \hline
Hardware Experiment 1    & 4     & Lissajous curve & Time varying speed   &              0.08  & 3     &      0.3         &       0.7        &      1          &    1   & 0.5       &    10     &    4    &   0.8       \\ \hline
Hardware Experiment 2    & 4     & Way-points & Time varying speed       &              0.15 & 3     &       0.3        &    0.7           &      1          &   1    &       0.5    &     10    &   2.5     &    0.8      \\ \hline
\end{tabular}
}\vspace{-0.4cm}
\label{tab_sim_expt}
\end{table*}

We  validated the proposed multi-agent convoy monitoring strategy in simulation and with real hardware experiments. We modelled the agents with unicycle kinematics and considered for different convoy paths with  smooth turns of different curvatures, and non-smooth paths with sharp turns in the MATLAB simulations. Here we do not consider the z-axis separation as it is a simulation to validate that the  cooperative scheme  achieves the desired formation. Since most aerial platforms can be approximated as unicycle agents, our strategy can be used with fixed-winged UAVs, helicopters and quadcopters. To demonstrate this, the proposed strategy has been simulated with a ROS-Gazebo based Software-In-The-Loop (SITL) simulator, where the aerial agents follow the 6-DOF dynamical model of a quadrotor. The presented strategy has also been implemented using the Bitcraze Crazyflie 2.0 quadrotors (equipped with the flowdeck sensor) in a motion capture arena for two different cases.  Table \ref{tab_sim_expt} summarises the cases considered for the simulations and the experiments.

\begin{figure}[!h]
\centering
\includegraphics[width=0.9\linewidth]{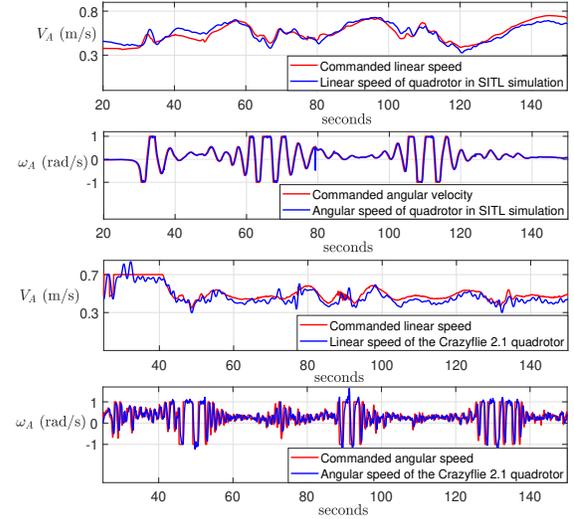}\\
\caption{PID tracking performance for quadrotors in the  SITL simulation and  the experiments}
\label{fig_pid_tracking}
\end{figure}

For the SITL simulation, the ground convoy follows waypoints on a road network like path with sharp turns to simulate motion in an urban environment. The SITL setup is the same as  in \cite{borkar2020reconfigurable}, wherein the 3DR Iris drone with the Pixhawk autopilot board is simulated. For the ground convoy we use a generic differential drive robot model. For both the SITL simulation and the experiments, the quadrotors have internal PID control loops which track  linear velocity commands along the global coordinate frame axes, i.e., $(V_x,V_y,V_z)$ and the angular velocity command $\omega_z$ about the vertical $z$ axes. We approximate the quadrotor as a unicycle agent, by mapping the linear speed $V_A$ and angular velocity $\omega_A$ given by \eqref{eqn_linear_speed_cmd} and \eqref{eqn_omg_a} as follows:
\begin{align}
V_x=V_A\cos(\psi_A),\ 
V_y=V_A\sin(\psi_A), \nonumber \\
V_z=k_z(z_{cmd}-z_A),\ 
\omega_z=\omega_A,
\end{align}
where $z_{cmd}$ is the commanded height of the quadrotor and $z_A$ is the height  of the quadrotor in the global reference frame. The PID loops in the flight controller firmware, for SITL simulation and the Crazyflie quadrotor, track the commanded linear speed and angular velocity satisfactorily as shown in Fig \ref{fig_pid_tracking}.
  
    \begin{figure}[!h]
\centering
\includegraphics[width=1\linewidth]{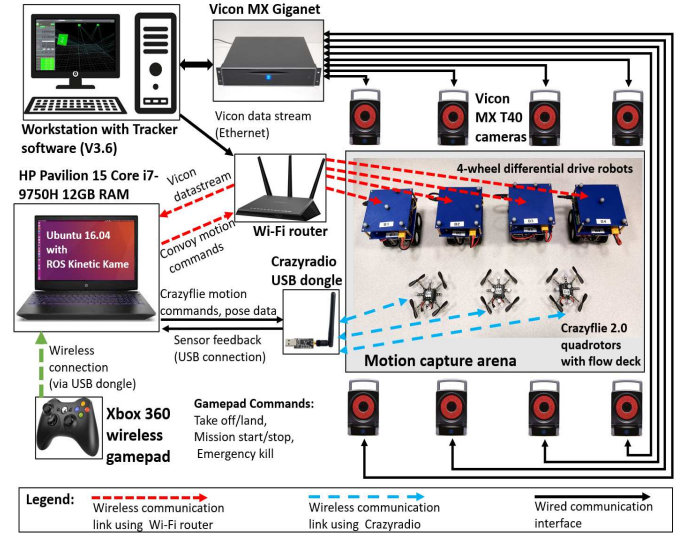}\\
\caption{Experiment Setup}
\label{fig_vicon_setup}
\end{figure}  
\begin{figure}[!h]
\centering
\includegraphics[width=1\linewidth]{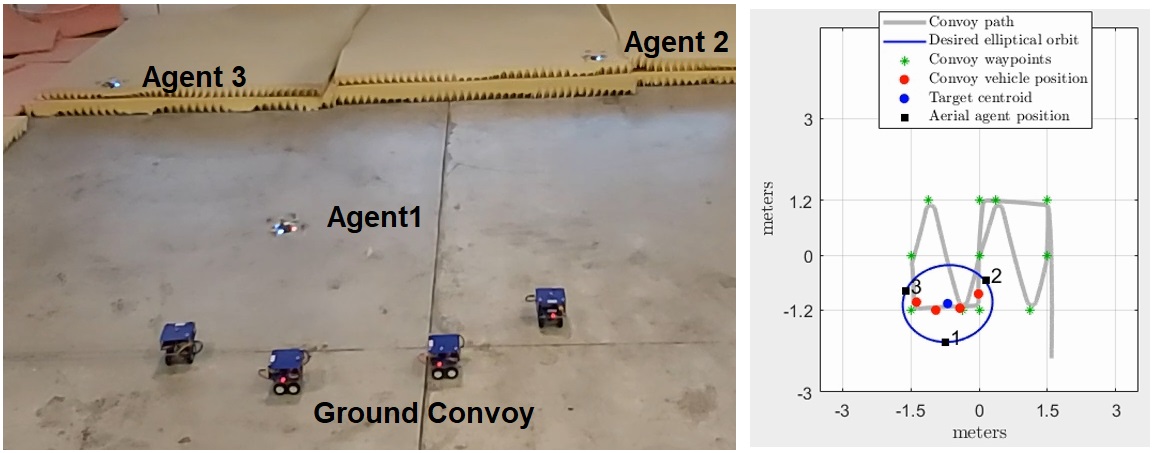}\\
\caption{(Left) Screenshot from experiment 2 showing the Crazyflie quadrotors monitoring the convoy of ground robots. (Right) Corresponding orthographic top view showing convoy path, and robot positions.}
\label{fig_expt2_path}
\end{figure} 
\begin{figure}[!h]
\centering
\includegraphics[width=1\linewidth]{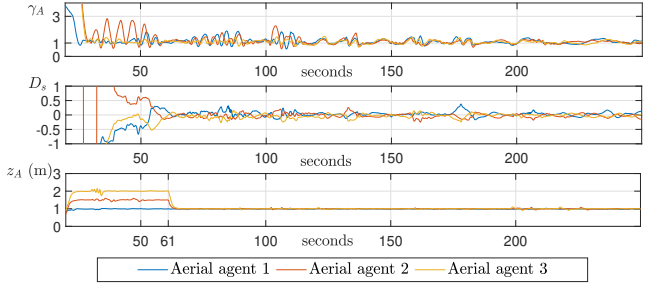}\\
\caption{Data plots for Experiment 2: For ideal orbit tracking  and formation control,  $\gamma_A=1$ and $D_s=0$.  The formation is achieved at $t=61$ sec, when the aerial agents are commanded to fly at the $z_{cmd}=1$ m.}
\label{fig_expt2_data}
\vspace{-0.6 cm}
\end{figure} 

   For the experiments we used small differential drive robots (designed in-house, refer Remark \ref{rem_vid}) as the convoy. The positions of the convoy robots and the quadrotors are available from the  motion capture system comprising of eight Vicon MX T40 cameras. The control implementation is decentralized in the sense that the control code for each convoy robot and each quadrotor is run as a separate ROS Node.
   
    The control node of each quadrotor receives its pose feedback and position coordinates of the convoy robots from the Vicon system. It cooperates with the control nodes for other quadrotors by broadcasting the  \textproc{Broadcast$\_$packet} described in Section \ref{sec_cooperation} on ROS topics. It also publishes motion commands to the corresponding Crazyflie quadrotor in the motion capture arena via the Crazyradio USB dongle-antenna. The ROS package used to facilitate this is the crazyflie\_ROS package \cite{crazyflieROS}. The control node for each convoy robot gets its pose information from the Vicon system and sends motion commands to the robot over a  Wi-Fi network. The Raspberry Pi 3B computer on each robot executes these commands. The experimental setup is illustrated in Fig.\ \ref{fig_vicon_setup}.

We discuss the plots of Experiment 2 given in Fig.\ \ref{fig_expt2_data}. The plots for the remaining simulation and experiment cases  can be found in the video, with web-link given in Remark \ref{rem_vid}. Ideally for perfect tracking of  elliptical orbit  $\gamma_A=1$ (given by \eqref{eqn_gamma}), and for exact formation separation, the separation error $D_s=0$ (given by \eqref{eqn_Ds}) for each agent. We thus consider these as the performance metrics for evaluating our strategy.
 Experiment 2 corresponds to the more challenging case where the convoy moves on a path having sharp turns as shown in Fig. \ref{fig_expt2_path}. Furthermore the ground convoy robots follow waypoints, while cooperating with each other and avoiding collisions. This causes them to move at different speeds, stop, reverse, etc. to ensure collision free motion.  Thus the aerial agents are tasked to track an orbit that is repeatedly changing its shape and orientation. From Fig.\ \ref{fig_expt2_data}, we see that after the formation is achieved,  despite some initial transients, the guidance law is able to drive the aerial agents to the desired orbit and thus maintain the value of $\gamma_A$ close to $1$, indicating that all aerial agents faithfully traverse the desired elliptical orbit as they monitor the moving convoy.  We see from Fig.\ \ref{fig_expt2_data} that  the error $D_s$ remains close to $0$, and significantly smaller than the desired parameter separation  $\Delta_s=\nicefrac{2\pi}{3}$, indicating that the aerial agents achieve and maintain the desired formation. The $z_A$ plot shows the initial altitude separation between the quadrotors for collision free motion on the ellipse, and equal altitudes are commanded when the $fl_{Hi}$ flag is set for each of the quadrotors according to Algorithm \ref{algo_coop}. 
The commanded $V_A$ and $\omega_A$ plots show that the linear speed and angular velocity constraints of the  aerial agents are always enforced. A similar explanation holds for the plots of the remaining simulations and experiment.
 \begin{remark}
\label{rem_vid}
 The videos of the simulations and experiments can be found at the web link:
\url{https://youtu.be/NjJ1F3E7UKs}
\end{remark}
 \section*{Conclusions}
We have proposed a multi-agent cooperative scheme for a team of UAVs to fly on time-varying elliptical orbits to monitor a ground convoy moving on an arbitrary continuous trajectory. The scheme computes a moving ellipse that  encompasses the  moving convoy. By using the proposed guidance law, the aerial agents converge to a trajectory that traverses the moving elliptical orbit repeatedly. To achieve and maintain a multi-agent formation on the elliptical orbit that divides its parametric length  into equal partitions, a cooperative control scheme is proposed to modulate the linear speed of the aerial agents. An algorithm is developed to ensure convergence to the desired formation and to avoid grouping of aerial agents at the same position due to input saturation. The proposed scheme maintains an approximately equi-parametric formation over the time varying elliptical orbit around the moving convoy. The proposed strategy has been validated by simulations in MATLAB and with a realistic simulation in a ROS-gazebo based SITL simulator for quadrotors. It has also been experimentally validated with quadrotors and ground robots in a motion capture arena.
This strategy provides sufficient autonomy to enable a single user to control a team of UAVs to monitor a moving convoy. Future directions include handling limited communication range and communication loss, and reconfigurable formations with in-mission addition and removal of agents.

 \section*{Acknowledgements}
The authors thank  Sri Theja Vuppala and  John M.\ Hart for their assistance with logistics for the experiments, which were implemented at the   Intelligent Robotics Laboratory, University of Illinois at Urbana Champaign. The work in this paper was funded in part by AFOSR\#FA9550-15-1-0146.
\vspace{-0.2 cm}


   \bibliographystyle{ieeetr}
\bibliography{References}

\begin{thebibliography}{10}

\bibitem{borkarborkarcdc}
A.~V. Borkar, V.~S. Borkar, and A.~Sinha, ``Vector field guidance for convoy
  monitoring using elliptical orbits,'' in {\em 56th IEEE Conference on
  Decision and Control (CDC)}, pp.~918--924, IEEE, 2017.

\bibitem{borkarborkarejcon}
A.~V. Borkar, V.~S. Borkar, and A.~Sinha, ``Aerial monitoring of slow moving
  convoys using elliptical orbits,'' {\em European Journal of Control},
  vol.~46, pp.~90--102, 2019.

\bibitem{targetsurvey}
C.~Robin and S.~Lacroix, ``Multi-robot target detection and tracking: taxonomy
  and survey,'' {\em Autonomous Robots}, vol.~40, no.~4, pp.~729--760, 2016.

\bibitem{beard_journal}
D.~R. Nelson, D.~B. Barber, T.~W. McLain, and R.~W. Beard, ``Vector field path
  following for miniature air vehicles,'' {\em IEEE Transactions on Robotics},
  vol.~23, no.~3, pp.~519--529, 2007.

\bibitem{vfield_cylinder}
D.~Lawrence, ``Lyapunov vector fields for {U}{A}{V} flock coordination.,'' in
  {\em AIAA 2nd Unmanned Unlimited Conference and Workshop, \& Exhibit},
  p.~6575, AIAA, 2003.

\bibitem{vfieldratnoo}
A.~A. Pothen and A.~Ratnoo, ``Curvature-constrained {L}yapunov vector field for
  standoff target tracking,'' {\em Journal of Guidance, Control, and Dynamics},
  vol.~40, no.~10, pp.~2729--2736, 2017.

\bibitem{frew_circle_standoff}
E.~Frew and D.~Lawrence, ``Cooperative stand-off tracking of moving targets by
  a team of autonomous aircraft,'' in {\em AIAA Guidance, Navigation, and
  Control Conference and Exhibit}, p.~6363, AIAA, 2005.

\bibitem{tsourdos_journal}
H.~Oh, S.~Kim, H.-s. Shin, and A.~Tsourdos, ``Coordinated standoff tracking of
  moving target groups using multiple {U}{A}{V}s,'' {\em IEEE Transactions on
  Aerospace and Electronic Systems}, vol.~51, no.~2, pp.~1501--1514, 2015.

\bibitem{frew_racetrack}
E.~W. Frew, D.~A. Lawrence, and S.~Morris, ``Coordinated standoff tracking of
  moving targets using {L}yapunov guidance vector fields,'' {\em AIAA Journal
  of Guidance, Control, and Dynamics}, vol.~31, no.~2, pp.~290--306, 2008.

\bibitem{frew_ellipse}
E.~W. Frew, ``Cooperative standoff tracking of uncertain moving targets using
  active robot networks,'' in {\em IEEE International Conference on Robotics
  and Automation (ICRA)}, pp.~3277--3282, IEEE, 2007.

\bibitem{galloway_cyc_pursuit}
K.~S. Galloway and B.~Dey, ``Station keeping through beacon-referenced cyclic
  pursuit,'' in {\em IEEE American Control Conference (ACC)}, pp.~4765--4770,
  IEEE, 2015.

\bibitem{ma_moving_cyc_pursuit}
L.~Ma and N.~Hovakimyan, ``Cooperative target tracking in balanced circular
  formation: Multiple {U}{A}{V}s tracking a ground vehicle,'' in {\em IEEE
  American Control Conference (ACC)}, pp.~5386--5391, IEEE, 2013.

\bibitem{ma_moving_var_rad}
L.~Ma and N.~Hovakimyan, ``Cooperative target tracking with time-varying
  formation radius,'' in {\em IEEE European Control Conference (ECC)},
  pp.~1699--1704, IEEE, 2015.

\bibitem{zhang_formation}
M.~Zhang and H.~H. Liu, ``Cooperative tracking of a moving target using
  multiple fixed-wing {U}{A}{V}s,'' {\em Journal of Intelligent \& Robotic
  Systems}, vol.~81, no.~3-4, pp.~505--529, 2016.

\bibitem{leonard_formation}
D.~Paley, N.~E. Leonard, and R.~Sepulchre, ``Collective motion: Bistability and
  trajectory tracking,'' in {\em 43rd IEEE Conference on Decision and Control
  (CDC)}, vol.~2, pp.~1932--1937, IEEE, 2004.

\bibitem{zhousatya}
B.~Zhou, H.~Satyavada, and S.~Baldi, ``Adaptive path following for unmanned
  aerial vehicles in time-varying unknown wind environments,'' in {\em IEEE
  American Control Conference (ACC)}, pp.~1127--1132, IEEE, 2017.

\bibitem{Kap}
Y.~A. Kapitanyuk, H.~Garcia~de Marina, A.~V. Proskurnikov, and M.~Cao,
  ``Guiding vector field algorithm for a moving path following problem,'' {\em
  20th IFAC World Congress}, vol.~50, no.~1, pp.~6983--6988, 2017.

\bibitem{muslimovadaptive}
T.~Z. Muslimov and R.~A. Munasypov, ``Adaptive decentralized flocking control
  of multi-uav circular formations based on vector fields and backstepping,''
  {\em ISA transactions}, pp.~S0019--0578.

\bibitem{ma2019ellipse}
L.~Ma, ``Cooperative target tracking in elliptical formation,'' in {\em 2019
  24th International Conference on Methods and Models in Automation and
  Robotics (MMAR)}, pp.~58--63, IEEE, 2019.

\bibitem{padmaji2012time}
V.~Padmaji, A.~Sinha, and H.~Arya, ``Time to capture target using cyclic
  pursuit strategy,'' in {\em Infotech@ Aerospace 2012}, p.~2526, 2012.

\bibitem{borkar2020reconfigurable}
A.~V. Borkar, S.~Hangal, H.~Arya, A.~Sinha, and L.~Vachhani, ``Reconfigurable
  formations of quadrotors on {L}issajous curves for surveillance
  applications,'' {\em European Journal of Control}, vol.~56, pp.~274--288,
  2020.

\bibitem{crazyflieROS}
W.~H{\"o}nig and N.~Ayanian, {\em Flying Multiple UAVs Using ROS}, pp.~83--118.
\newblock Springer International Publishing, 2017.

\end{thebibliography}
\end{document}